\documentclass[runningheads]{llncs}
\let\llncssubparagraph\subparagraph
\let\subparagraph\paragraph

\usepackage[a4paper,total={170mm,257mm},left=20mm,top=20mm,]{geometry}
\usepackage{float}  
\usepackage{hyphenat}
\usepackage{graphicx}
\usepackage{booktabs} 
\usepackage{subfigure}
\usepackage{caption}
\usepackage{xcolor}
\usepackage{enumitem}
\usepackage{hyperref}
\usepackage{array}
\usepackage{booktabs}
\usepackage{tikz}
\usepackage{pbox}
\usepackage{bm}
\usepackage{titlesec}
\let\subparagraph\llncssubparagraph
\usepackage{lipsum}

\begin{document}
\title{FMD-cGAN: Fast Motion Deblurring using Conditional Generative Adversarial Networks}
%
%
\author{Jatin Kumar\inst{1}\orcidID{0000-0002-3418-1773} \and
Indra Deep Mastan\inst{2}\orcidID{0000-0001-5033-9561} \and
Shanmuganathan Raman\inst{1}\orcidID{0000-0003-2718-7891}}
\authorrunning{J. Kumar et al.}
%
\institute{IIT Gandhinagar, India \\ \email{\{kumar\_jatin\}@alumni.iitgn.ac.in, \{shanmuga\}@iitgn.ac.in } \and LNMIIT Jaipur, India \email{\{indradeep.mastan\}@lnmiit.ac.in }}
\maketitle              
\begin{abstract}
In this paper, we present a Fast Motion Deblurring-Conditional Generative Adversarial Network (FMD-cGAN) that helps in blind motion deblurring of a single image. FMD-cGAN delivers impressive structural similarity and visual appearance after deblurring an image. Like other deep neural network architectures, GANs also suffer from large model size (parameters) and computations. It is not easy to deploy the model on resource constraint devices such as mobile and robotics. With the help of MobileNet{~\color{green}~\cite{mobilenetv1}} based architecture that consists of depthwise separable convolution, we reduce the model size and inference time, without losing the quality of the images. More specifically, we reduce the model size by 3-60x compare to the nearest competitor. The resulting compressed Deblurring cGAN faster than its closest competitors and even qualitative and quantitative results outperform various recently proposed state-of-the-art blind motion deblurring models. We can also use our model for real-time image deblurring tasks. The current experiment on the standard datasets shows the effectiveness of the proposed method.

\end{abstract}
\section{Introduction}
\label{sec:IntroductionSection}
\begin{figure}[!htb]
\centering
\subfigure[Blur Image]{
  \includegraphics[width=0.36\linewidth,height=35mm]{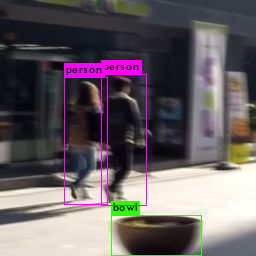}
}
\subfigure[Sharp Image]{
  \includegraphics[width=0.36\linewidth,height=35mm]{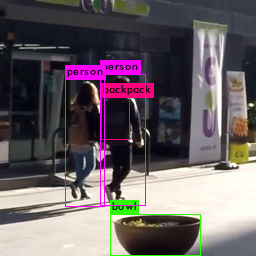}
}
\setlength{\belowcaptionskip}{-15pt}
\caption{The figure shows object detection on images becomes easy after deblurring using FMD-cGAN.YOLO{\color{green}\cite{yolo2015detection}} object detection on the (a) blurry picture and on the (b) sharp picture from the GoPro dataset{\color{green}\cite{deblurgan}}}
\label{fig:object_detection}
\vspace{-1mm}
\end{figure}

Image degradation by motion blur generally occurs due to movement during the capture process from the camera or capturing using lightweight devices such as mobile phones and low intensity during camera exposure. 
Blur in the images degrades the perceptual quality. For example, blur distorts the object's structure  (Fig. {\color{red} \ref{fig:object_detection}}).

Image Deblurring is a method to remove the blurring artifacts and distortion from a blurry image. Human vision can easily understand the blur in the image. However, it is challenging to create metrics that can estimate the blur present in the image. Image degradation model using non-uniform blur kernel {\color{green} \cite{blur2017flow,nonuniform2015blur}} is given in Eq.~{\color{red}\ref{eq:Image deblur}}.
\begin{equation}
I_B = K(M) * I_S + N
\label{eq:Image deblur}
\end{equation}
where, $I_B$ denotes a blurred image, K(M) denotes unknown blur kernels depending on M's motion field. $I_S$ denotes a latent sharp image, $*$ denotes a convolution operation, and N denotes the noise. As an inverse problem, we retrieve sharp image $I_S$ from blur image $I_B$ during the deblurring process. The deblurring problem generally classified as non-blind deblurring {\color{green} \cite{nonblind2006deblurring}} and blind deblurring {\color{green} \cite{blind2010deblur,blind2011deblur}}, according to knowledge of blur kernel $K(M)$ is known or not. 

Our work aims at a single image blind motion deblurring task using deep-learning. The deep-learning methods are effective in performing various computer vision tasks such as object removal {\color{green} \cite{objrem2018mulmed,srobust}}, style transfer {\color{green} \cite{deep2017styletransf}}, and image restoration {\color{green} \cite{sr_photo_reliastic,deblurgan,deblur_dynamicscene}}. More specifically, convolution neural networks (CNNs) based approaches for image restoration tasks are increasing, e.g., image denoising {\color{green} \cite{gaussiandenoise}}, super-resolution {\color{green} \cite{sr_photo_reliastic}}, and  deblurring {\color{green} \cite{deblurgan,deblur_dynamicscene}}.

The applications of Generative Adversarial Networks (GANs) {\color{green} \cite{gan2014ian}} are increasing immensely, particularly image-to-image conversion GANs {\color{green} \cite{pix2pix}} have been successfully used on image enhancement, image synthesis, image editing and style transfer. Image deblurring could be formulated as an image-to-image translation task.  Generally, applications that interact with humans (e.g., Object Detection) require to be faster and lightweight for a better experience. Image deblurring could be useful pre-processing steps of other computer vision tasks such as Object Detection (Fig. {\color{red} \ref{fig:object_detection}}).  

\begin{figure*}
\centering
\begin{minipage}{0.3\linewidth}
  \includegraphics[width=\linewidth,height=30mm]{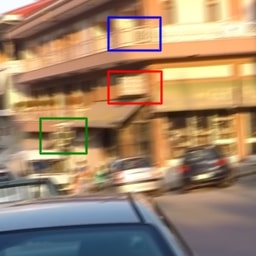}
\end{minipage}\hspace*{5pt}
\begin{minipage}{0.3\linewidth}
  \includegraphics[width=\linewidth,height=30mm]{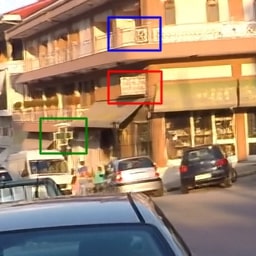}
\end{minipage}\hspace*{5pt}
\begin{minipage}{0.3\linewidth}
  \includegraphics[width=\linewidth,height=30mm]{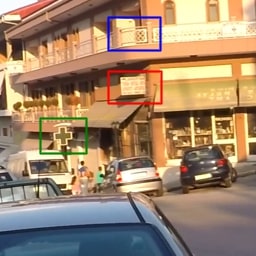}
\end{minipage} \\%
\begin{minipage}{0.3\linewidth}
  \includegraphics[width=\linewidth,height=30mm]{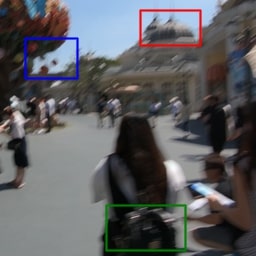}\\ \hspace*{1cm} (a) Corrupted Image
\end{minipage}\hspace*{5pt}
\begin{minipage}{0.3\linewidth}
  \includegraphics[width=\linewidth,height=30mm]{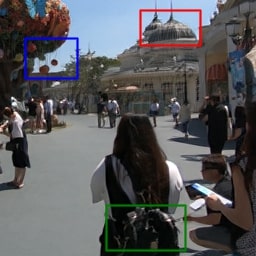}\\ \hspace*{1cm} (b) FMD-cGAN (ours)
\end{minipage}\hspace*{5pt}
\begin{minipage}{0.3\linewidth}
  \includegraphics[width=\linewidth,height=30mm]{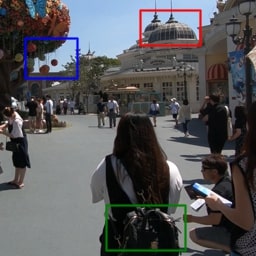}\\ \hspace*{1cm} (c) Original Image
\end{minipage}%
 \caption{First-row images are from the GoPro dataset {\color{green} \cite{multi_scale2017}}, and second-row images are from the REDS dataset {\color{green} \cite{reds}} processed by Fast Deblurring cGAN.}
\label{fig:gopro_result}
\vspace{-8mm}
\end{figure*}

In this paper, we propose a Fast Motion Deblurring conditional Generative Adversarial Network architecture (FMD-cGAN). Our FMD-cGAN architecture is based on conditional GANs {\color{green} \cite{conditional2014gan}} and the resnet network architecture {\color{green} \cite{deep_res}} (Fig. {\color{red} \ref{fig:generator architecture}}). We also used depthwise separable convolution (Fig. {\color{red} \ref{fig:modified_resnet_block}}) inspired from MobileNet to improve efficiency. A MobileNet network {\color{green} \cite{mobilenetv1}} has fewer Multiplications and Additions (smaller complexity) operations, and fewer parameters (smaller model size) compare to the same network with regular convolution operation. 

Unlike other GAN frameworks, where we give the sharp image (real example) and output image from generator network (fake example) as the inputs into  Discriminator network {\color{green} \cite{pix2pix,conditional2014gan}}, we train our Discriminator (Fig. {\color{red} \ref{fig:Discriminator architecture}}) by providing input as combining blurred image with the output image from the generator network (or blurred image with sharp image).

Different from previous work, we propose to use Hinge loss {\color{green} \cite{geometric_gan}} and Perceptual loss {\color{green} \cite{perceptualloss}} to improve the quality of the output image. Hinge loss improves the fidelity and diversity of the generated images {\color{green} \cite{autogan}}. Using the Hinge loss in our FMD-cGAN allows building lightweight neural network architectures for the single image motion deblurring task compared to standard Deep ResNet architectures. The Perceptual loss {\color{green} \cite{perceptualloss}} is used as content loss to generate photo-realistic images in our GAN framework. \vspace*{4pt}

\noindent \textbf{Contributions:} The major contributions are summarized as below. 
\begin{itemize}
    \item[$\bullet$] We propose a faster and light-weight conditional GAN architecture (FMD-cGAN) for blind motion deblurring tasks.      We show that FMD-cGAN (ours) is efficient with lesser inference time than DeblurGAN {\color{green} \cite{deblurgan}}, DeblurGANv2 {\color{green} \cite{deblur_v2}}, and DeepDeblur {\color{green} \cite{multi_scale2017}} models (Table {\color{red} \ref{tab:Performance and efficiency comparison on the GoPro test dataset}}).
    \item[$\bullet$] We have performed extensive experiments on GoPro dataset and REDS dataset (Sec.~{\color{red}\ref{sec:experimentalResults}}). The results shows that our FMD-cGAN outputs images with good visual quality and structure similarity (Fig~{\color{red} \ref{tbl:comparison on the REDS dataset}}, Fig.~{\color{red} \ref{tbl:comparison on the GoPro dataset}}, and Table~{\color{red} \ref{tab:Performance and efficiency comparison on the REDS test dataset}}). 
    \item[$\bullet$] We also provide two variants (WILD and Comb) of FMD-cGAN to show that image deblurring task could be improved by pre-training network (Table {\color{red} \ref{tab:Performance and efficiency comparison on the GoPro test dataset}} and {Sec. \color{red} \ref{sec:training_details}}). 
    \item[$\bullet$] We have also performed ablation study to illustrate that our network design choices improves the deblurring performance (Sec. {\color{red} \ref{sec:ablation_study}}).
\end{itemize}

\section{Background}
\subsection{Image Deblurring}
Images can have different types of blur problems, such as motion blur, defocus blur, and handshake blur. We have described that image deblurring is classified into two types: Non-blind image deblurring and Blind image deblurring (Sec.~{\color{red} \ref{sec:IntroductionSection}}).


Non-blind deblurring is an ill-posed problem. The noise inverse process is unstable; a small quantity of noise can cause critical distortions. Most of the earlier works {\color{green} \cite{cv_richard,william_hadley,wiener_filter}} aims to perform non-blind deblurring task by assuming that blur kernels $K(M)$ are known. Blind deblurring techniques for a single image, which use Deep-learning based approaches, are observed to be effective in single image deblurring tasks {\color{green} \cite{SRN,multi_scale2017}} because most of the kernel-based methods are not sufficient to model the real world blur {\color{green} \cite{raw2020deblur}}. The task is to estimates both the sharp image $I_S$ and the blur kernel $K(M)$ for image restoration. There are also classical approaches such as low-rank prior {\color{green}\cite{lowrank_prior}} and dark channel prior {\color{green}\cite{darkchannel_prior}} that are useful for deblurring, but they also have shortcomings. 

\subsection{Generative Adversarial Networks}
Generative Adversarial Network (GAN) was initially developed and introduced by Ian Goodfellow and his fellow workers in 2014 {\color{green} \cite{gan2014ian}}.  GAN framework includes two competing network architectures: a generator network $G$ and a discriminator network $D$.
\noindent Generator ($G$) task is to generate fake samples similar to input by capturing the input data distribution, and on the opposite side, the Discriminator ($D$) aims to differentiate between the fake and real samples; and pass this information to the $G$ so that $G$ can learn.  Generator $G$ and Discriminator $D$ follows the minimax objective defined as follows. 
\begin{equation}
\min_G\max_D V(D,G) = E_{x\sim p_{data}(x)}[log(D(x))]+E_{z\sim p_{z}(z)}[log(1 - D(G(z))]
\label{eq:gan minimax}
\end{equation}
Here, in Eq.~{\color{red}\ref{eq:gan minimax}}, the generator $G$ aims to minimize the value function $V$, and  the discriminator $D$ tries to maximize the value function $V$. Moreover, the generator $G$ faces problems such as mode collapse and gradient diminishing (e.g., Vanilla GAN). \\

\noindent \textbf{WGAN and WGAN-GP:} To deal with mode collapse and gradient diminishing, WGAN method {\color{green} \cite{wgan}} uses Earth-Mover (Wasserstein-1) distance in the loss function. In this implementation, the discriminator output layer is a linear one, not sigmoid (discriminator output's a real value). WGAN {\color{green} \cite{wgan}} performs weight clipping $[{-c},\ c]$ to enforce the Lipschitz constraint on the critic (i.e., discriminator). This method faces the issue of gradient explosion/vanishing without proper value of weight clipping parameter $c$. WGAN with Gradient penalty (WGAN-GP) {\color{green} \cite{gradientpenalty}} resolve above issues with WGAN {\color{green} \cite{wgan}}. WGAN-GP enforces a penalty on the gradient norm for random samples $\tilde{x} \sim P_{\tilde{x}}$.
The objective function of WGAN-GP is as below.

\begin{equation}
V(D,G) = \min_G\max_D E_{\tilde{x} \sim p_{g}}[D(\tilde{x})] - E_{x \sim p_{r}} [D(x)] +\lambda E_{\tilde{x} \sim P_{\tilde{x}}} [(||{\nabla_{\tilde{x}}D(\tilde{x})}||_2 - 1)^2]
\label{eq:gradient panelty}
\end{equation}

\noindent WGAN-GP {\color{green} \cite{gradientpenalty}} makes the WGAN {\color{green} \cite{wgan}} training more stable and does not require hyperparameter tuning. The DeblurGAN {\color{green} \cite{deblurgan}} used WGAN-GP method (Eq. {\color{red}\ref{eq:gradient panelty}}) for single image blind motion deblurring. \\

\noindent \textbf{Hinge Loss:} In our method, we used Hinge loss {\color{green} \cite{geometric_gan,sa_gan}} which is giving better result as compared to WGAN-GP {\color{green} \cite{gradientpenalty}} based deblurring method. Hinge loss output also a real value. 
Generator loss $L_G$ and Discriminator loss $L_D$ in the presence of Hinge loss is defined as follows.
\begin{equation}
L_D ={} -E_{(x,y) \sim p_{data}}[min(0, -1+D(x,y))] -E_{z \sim p_{z}, y \sim p_{data}}[min(0,-1-D(G(z),y))]
\label{eq:hinge discriminator}
\end{equation}

\begin{equation}
L_G =  - E_{z \sim p_{z}, y \sim p_{data}}D(G(z),y)
\label{eq:hinge generator}
\end{equation}
\noindent Here, $D$ tries that a real image will get a large value, and a fake or generated image will get a small value.

\begin{figure}
\begin{minipage}{0.45\textwidth}
    \includegraphics[width=\linewidth,height=45mm]{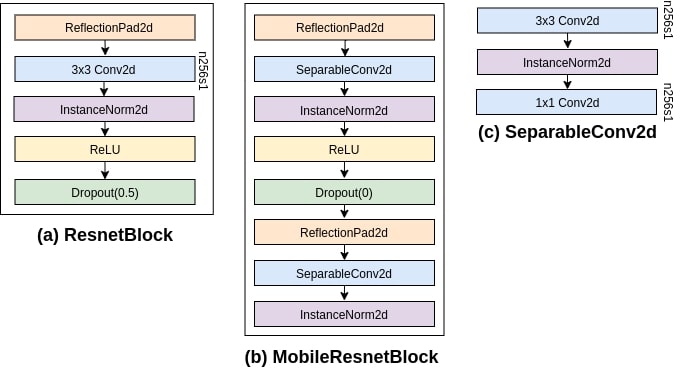}
    \caption{Modified Resnet Block}
    \label{fig:modified_resnet_block}
\end{minipage}\hspace*{0.75cm}%
\begin{minipage}{0.5\textwidth}
    \includegraphics[width=\linewidth, height=35mm]{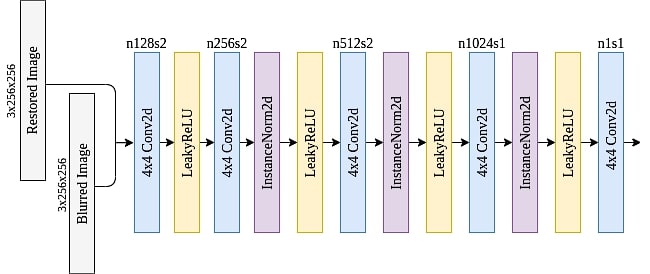}
    \caption{The figure shows the architecture of the critic network (Discriminator). }
    \label{fig:Discriminator architecture}
\end{minipage}
\vspace{-6mm}
\end{figure}

\section{Related Works}
The deep learning-based methods attempt to estimate the motion blur in the degraded image and use this blurring information to restore the sharp image {\color{green} \cite{nonunblrrem}}. The methods which use the multi-scale framework {\color{green} \cite{multi_scale2017}} to recover the deblurred image are computationally expensive. The use of GANs also increasing in blind kernel free single image deblurring tasks such as Ramakrishnan et al. {\color{green} \cite{ramakrishnan}} used image translation framework {\color{green} \cite{pix2pix}} and densely connected convolution network {\color{green} \cite{densenetwork}}. The methods above performs  image-deblurring task, when input image may have blur due to multiple sources. Kupyn et al. {\color{green} \cite{deblurgan}} proposed the DeblurGAN method, which uses the Wasserstein GAN {\color{green} \cite{wgan}} with gradient penalty {\color{green} \cite{gradientpenalty}} and the Perceptual loss {\color{green} \cite{perceptualloss}}. Kupyn et al. {\color{green} \cite{deblur_v2}} proposed a new method DeblurGAN-v2, which is faster and has better results than the previously proposed method; this method uses the feature pyramid network {\color{green} \cite{pyramid_network}} in the generator. A study of various single image blind deblurring methods is provided in {\color{green} \cite{cvpr16_deblur_study}}.

\section{Our Method}
In our proposed method, the blur kernel knowledge is not present, and from a given blur image $I_B$ as an input, our purpose is to develop a sharp image $I_S$ from $I_B$. For the deblurring task, we train a Generator network denoted by $G_{\theta_G}$. During the training period, along with Generator, there is one another CNN also present $D_{\theta_D}$ referred to as the critic network (i.e., Discriminator). The Generator $G_{\theta_G}$ and the Discriminator $D_{\theta_D}$ are trained in an adversarial manner. In what follows, we describe the network architecture and the loss functions for our method. 

\subsection{Network Architecture}
The generator network, a chief component of proposed model, is a transformed version of residual network architecture {\color{green} \cite{deep_res}} (Sec.{\color{red}~\ref{ssec:generatorArchitecture})}. The discriminator architecture, which helps to learn the Generator, is a transformed version of Markovian Discriminator (PatchGAN) {\color{green} \cite{pix2pix}} (Sec.{\color{red}~\ref{ssec:discriminatorArchitecture})}. The residual network architecture helps us to build deeper CNN architectures. Also, this architecture is effective because we want our network to learn only the difference between pairs of sharp and blur images as they are almost alike in values. 
 
 We used the depthwise separable convolution in place of the standard convolution layer to reduce the inference time and model size {\color{green} \cite{mobilenetv1}}. Generator aims to generate sharp images given the blurred images as input. Note that generated images need to be realistic so that the Discriminator thinks that generated images are from the real data distribution. In this way, the Generator helps to generate a visually attractive sharp image from an input blurred image. Discriminator goal is to classify if the input is from the real data distribution or output from the generator. Discriminator accomplish this by analyzing the patches in the input image for making a decision. The changes which we made in the resnet block displayed in Fig. {\color{red} \ref{fig:modified_resnet_block}}, we convert structure (a) into structure (b).

  \begin{figure*}[h]
    \centering
    \includegraphics[width=\linewidth]{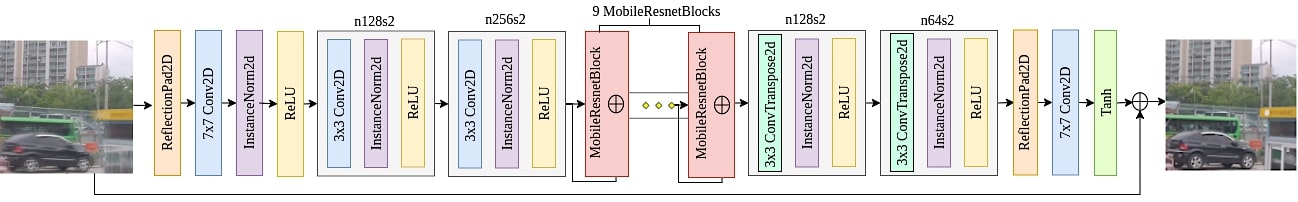}
    \caption{The figure shows the generator architecture of our Fast Motion Deblurring-cGAN. Given a blurred image as input, Generator outputs a realistic-looking sharp image as output.}
    \label{fig:generator architecture}
    \vspace{-8mm}
\end{figure*}

\subsubsection{Generator Architecture.}
\label{ssec:generatorArchitecture}
The Generator's CNN architecture is displayed in Fig. {\color{red} \ref{fig:generator architecture}}. This architecture is alike to the style transfer architecture which is proposed by Johnson et al. {\color{green} \cite{perceptualloss}}. The generator network has two strided convolution blocks in begining with stride 2, nine depthwise separable convolutions based residual blocks (MobileResnet Block) {\color{green} \cite{mobilenetv1,deep_res}}, and two transposed convolution blocks, and the global skip connection.
In our architecture, most of the computation is done by MobileResNet-Block. Therefore, we use depthwise separable convolution here to reduce computation cost without affecting accuracy.
 
Every convolution and transposed convolution layer have an instance normalization layer {\color{green} \cite{inst_norm}} and a ReLU activation layer {\color{green} \cite{dl2019relu}} behind it. Each Mobile Resnet block consists of two depthwise separable convolutions {\color{green} \cite{mobilenetv1}}, a dropout layer {\color{green} \cite{dropout}}, two instance normalization layers after each separable convolution block, and a ReLU activation layer. In each mobile resnet block, after the first depth-wise separable convolution layer, a dropout regularization layer with a probability of zero is added. Furthermore, we add a global skip connection in the model, also referred to as ResOut. 

When we use many convolution layers, it will become difficult to generalize over first-level features, deep generative CNNs often unintentionally memorize high-level representations of edges. The network will be unable to retrieve sharp boundaries at proper positions from the blur photos as a result of this. We combine the head and tail of the network. Since the gradients value now can reach from the tail straight to the beginning and affect the update in the lower layers, generation efficiency improves significantly {\color{green} \cite{idnmap}}. In the blurred image $I_B$, CNN learns residual correction $I_R$, so the resulting sharp image is $I_S = I_B + I_R$. From experiments, we come to know that such formulation improves the training time, and generalizes the resulting model better.

\subsubsection{Discriminator Architecture.}
\label{ssec:discriminatorArchitecture}
In our model, we create a critic network $D_{\theta_D}$ also refer to as Discriminator. $D_{\theta_D}$ guides Generator network $G_{\theta_G}$ to generate sharp images by giving feedback on the input is from real data distribution or generator output. The architecture of Discriminator network is shown in Fig. {\color{red} \ref{fig:Discriminator architecture}}. We avoid high-depth Discriminator network as it's goal is to perform the classification task unlike image synthesis task of Generator network. In our FMD-cGAN framework, the Discriminator network is similar to the Markovian patch discriminator, also refer to as PatchGAN {\color{green} \cite{pix2pix}}. Except for the last convolutional layer, InstanceNorm layer and LeakyReLU with a value of 0.2, follow all convolutional layers of the network. This architecture looks for explicit structural characteristics at many local patches. It also ensures that the generated raw images have a rich color.

\begin{table*}[htb]
\fontsize{8.5}{10}\selectfont \renewcommand{\arraystretch}{1.2}
\begin{center}
\begin{tabular}{|c|c|c|c|c|c|c|}
\hline
\textbf{Method} & \textbf{PSNR} & \textbf{SSIM} & \textbf{Time (GPU)} & \textbf{Time (CPU)} & \textbf{\#Parameters} & \textbf{MACs} \\
\hline
Sun et al. {\color{green} \cite{nonuniform2015blur}} & 24.64 & 0.842 & N/A & N/A & N/A & N/A\\
\hline
Xu et al. {\color{green} \cite{unnaturalsparse}} & 25.1 & 0.89 & N/A & N/A & N/A & N/A\\
\hline
DeepFilter {\color{green} \cite{ramakrishnan}} & 28.94 & 0.922 & 0.3 sec & 3.09 sec & 3.20M & N/A\\
\hline
$DeblurGAN_{WILD}$ {\color{green} \cite{deblurgan}} & 27.2 & 0.954 & 0.45 sec & 3.36 sec & 6.06M & 35.07G\\
$DeblurGAN_{Comb}$  & 28.7 & 0.958 & & & & \\
\hline
$DeblurGANv2_{Resnetv2}$ {\color{green} \cite{deblur_v2}} & 29.55 & 0.934 & 0.14 sec & 3.67 sec & 66.594M & 274.20G\\
$DeblurGANv2_{Mobnetv2}$ & 28.17 & 0.925 & 0.04 sec & 1.23 sec & 3.12M & 39.05G \\
\hline
SRN {\color{green} \cite{SRN}} & 30.10 & 0.932 & 1.6 sec & 28.85 sec & 6.95M & N/A\\
\hline
DeepDeblur {\color{green} \cite{multi_scale2017,github_deepdeblur}} & \textbf{30.40} & 0.901 & 2.93 sec & 56.76 sec & 11.72M & 4727.22G\\
\hline
FMD-cGAN$_{WILD}$  & 28.33 & 0.962 & \textbf{0.01 sec} & \textbf{0.28 sec} & \textbf{1.98M} & \textbf{18.36G} \\
FMD-cGAN$_{Comb}$  & 29.675 & \textbf{0.971} & & & & \\
\hline
\end{tabular}
\end{center}
\caption{The table shows the results on GoPro test dataset. Here, FMD-cGAN$_{WILD}$ and FMD-cGAN$_{Comb}$ are our methods (Sec.~\ref{sec:training_details}). It could be observed that our frameworks achieves good quantitative performance.  }
\label{tab:Performance and efficiency comparison on the GoPro test dataset}
\vspace{-12mm}
\end{table*}

\subsection{Loss Functions}
The total loss function for FMD-cGAN deblurring framework is the mixture of adversarial loss and content loss.
\begin{equation}
    L_{total} = L_{GAN} + \lambda \cdot L_X
    \label{eq:total loss}
\end{equation}
In Eq.~{\color{red}\ref{eq:total loss}}, $L_{GAN}$ represents the advesarial loss (Sec.{\color{red} \ref{sec:adversarial_loss}}), $L_X$ represents the content loss (Sec.{\color{red} \ref{sec:content_loss}}) and $\lambda$ represents the hyperparameter which controls the effect of $L_X$. The value of $\lambda$ is equal to 100 in the current experiment.
\\

\subsubsection{Adversarial Loss.}
\label{sec:adversarial_loss}
To train a learning-based image restoration network, we need to compare the difference between the restored and the original images during the training stage. Many image restoration works are using an adversarial-based network to generate sharp images {\color{green} \cite{sr_photo_reliastic,deblur_dynamicscene}}. During the training stage, the adversarial loss after pooling with other losses helps to determine how good the Generator is working against the Discriminator {\color{green} \cite{multi_scale2017}}. Initial works based on conditional GANs use the objective function of the vanilla GAN as the loss function {\color{green} \cite{sr_photo_reliastic}}. Lately, least-square GAN {\color{green} \cite{lsgan}} was observed to be better balanced and produce the good quality desired outputs. We apply Hinge loss {\color{green} \cite{geometric_gan}} (Eq. {\color{red} \ref{eq:hinge discriminator}} and Eq. {\color{red}\ref{eq:hinge generator}}) in our model to provide good results with the generator architecture {\color{green} \cite{autogan}}. Generator loss ($L_G$) and Discriminator loss ($L_D$) are computed as follows (Eq.~{\color{red}\ref{eq:hinge generator loss}} and Eq.~{\color{red}\ref{eq:hinge discriminator loss}}).
\begin{equation}
    L_G = - \sum\limits_{n=1}^N D_{\theta_D}(G_{\theta_G}(I^B))
    \label{eq:hinge generator loss}
\end{equation}
\begin{equation}
    L_D = - \sum\limits_{n=1}^N min(0, D_{\theta_D}(I^S)-1) - \sum\limits_{n=1}^N min(0, -D_{\theta_D}(G_{\theta_G}(I^B))-1)
    \label{eq:hinge discriminator loss}
\end{equation}

If we do not use adversarial loss in our network, it still converges. However, the output images will be dull with not many sharp edges, and these output images are still blurry because the blur at edges and corners is still intact. If we only use adversarial loss in our network, edges are retained in images, and more practical color assignment happens. However, it has two issues:  still, it has no idea about the structure, and Generator is working according to the guidance provided by Discriminator based on the generated image. We remove these issues with the adversarial loss by combining adding with the Perceptual loss.

\subsubsection{Content loss.}
\label{sec:content_loss}
Generally, there are two choices for the pixel-based content loss: (a) L1 or MAE loss and (b) L2 or MSE loss. Moreover, above loss functions may  produce blurry artifacts on the generated image due to the average of pixels {\color{green} \cite{sr_photo_reliastic}}. Due to this issue, we used Perceptual loss {\color{green} \cite{perceptualloss}} function for content loss. Unlike L2 Loss,  Perceptual compares the difference between  CNN feature maps of the restored image and the original image. This loss function puts structural knowledge into the Generator, which helps it against the patch-wise decision of the Markovian Discriminator. The equation of the Perceptual loss is as follows:
\begin{equation}
    L_X = \frac{1}{W_{i,j}H_{i,j}} \sum\limits_{x=1}^{W_{i,j}} \sum\limits_{y=1}^{H_{i,j}} (\phi_{i,j}(I^S)_{x,y} - \phi_{i,j}(G_{\theta_G}(I^B))_{x,y})^2
    \label{eq:Perceptual loss}
\end{equation}
Where $W_{i,j}$ and $H_{i,j}$ are the width and height of the $(i,j)^{th}$ ReLU layer of the \textbf{VGG-16} network \cite{vgg_19}, here i and j denote ${j^{th}}$ convolution (\,after activation)\, before the $i^{th}$ max-pooling layer. $\phi_{i,j}$ denotes the feature map. In our current method, we use the output of activations from $VGG_{3,3}$ convolutional layer. The output from activations of the end layers of the network represents more features information {\color{green} \cite{sr_photo_reliastic,vu_cnn}}. The Perceptual loss helps to restore the general content {\color{green} \cite{pix2pix,sr_photo_reliastic}}; on the other side adversarial loss helps to restore texture details. If we do not use the Perceptual loss in our network or use simple MSE based loss on pixels, the network will not converge to a good state.


\subsection{Training Datasets}
\subsubsection{GoPro Dataset.} The images of the GoPro dataset {\color{green} \cite{multi_scale2017}} are generated using the GoPro Hero 4 camera. The camera captures 240 frames per second video sequences. The blurred images are captured by averaging consecutive short-exposure frames. It is the most commonly used benchmark dataset in motion deblurring tasks, containing 3214 pairs of blur and sharp images. We use 1111 pairs of images for testing purposes and the remaining 2103 pairs of images for training {\color{green} \cite{multi_scale2017}}.
 
\subsubsection{REDS Dataset.} The Realistic and Dynamic Scenes dataset {\color{green} \cite{reds}} was designed for video deblurring and super-resolution, but it is also helpful in the image deblurring. The dataset comprises 300 video sequences having a resolution of 720×1280. Here, the training set contains 240 videos, the validation set contains 30 videos, and the testing set contains 30 videos. Each video has 100 frames. REDS dataset is generated from 120 fps videos, synthesizing blurry frames by merging subsequent frames. We have 240*100 pairs of blur and sharp images for training, 30*100 pairs of blur and sharp images for testing.
     

\begin{table*}
        \centering
        \begin{tabular}{cccc}
           \toprule
            \textbf{Blurry} & \textbf{DeepDeblur} {\color{green} \cite{multi_scale2017}} & \textbf{FMD-cGAN(Ours)} & \textbf{Sharp} \\
            \midrule
            \includegraphics[width=0.24\linewidth,height=30mm]{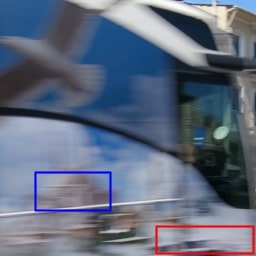} &  \includegraphics[width=0.24\linewidth,height=30mm]{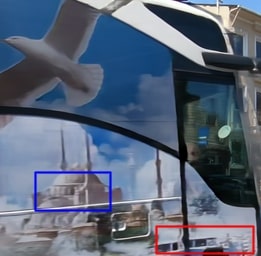} &
            \includegraphics[width=0.24\linewidth,height=30mm]{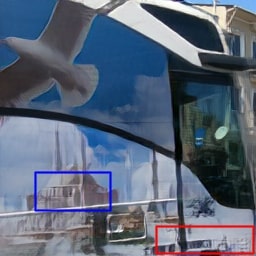} &
            \includegraphics[width=0.24\linewidth,height=30mm]{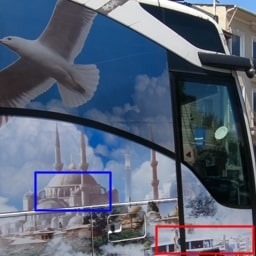} \\
            \includegraphics[width=0.24\linewidth,height=30mm]{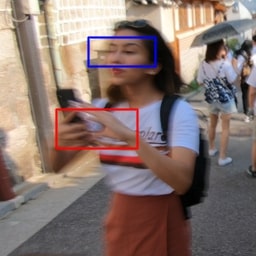} &  \includegraphics[width=0.24\linewidth,height=30mm]{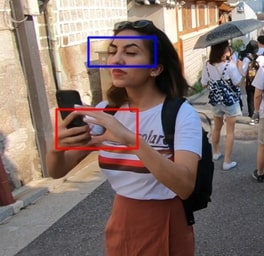} &
            \includegraphics[width=0.24\linewidth,height=30mm]{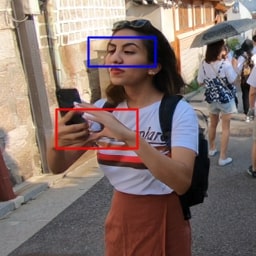} &
            \includegraphics[width=0.24\linewidth,height=30mm]{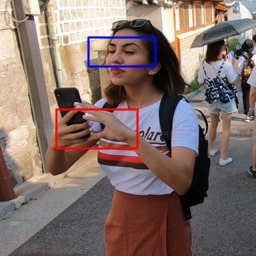} \\
            \includegraphics[width=0.24\linewidth,height=30mm]{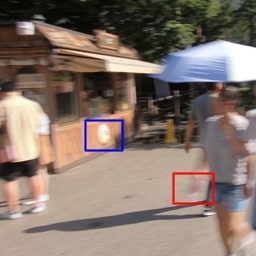} &  \includegraphics[width=0.24\linewidth,height=30mm]{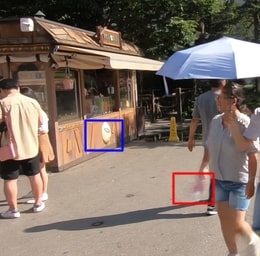} &
            \includegraphics[width=0.24\linewidth,height=30mm]{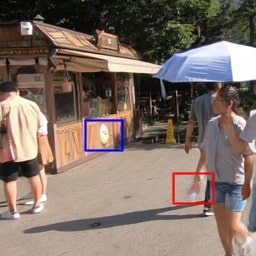} &
            \includegraphics[width=0.24\linewidth,height=30mm]{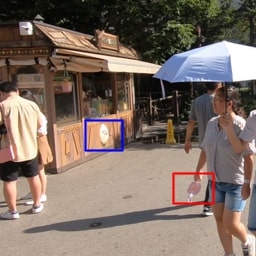} \\
            \bottomrule
        \end{tabular}
        \captionof{figure}{The figure shows visual comparison on the REDS dataset (images are best viewed after zooming).}
        \label{tbl:comparison on the REDS dataset}
        \vspace{-6mm}
\end{table*}

\begin{table}[!ht]
\renewcommand{\arraystretch}{1.2}
\begin{minipage}{0.45\textwidth}
\begin{center}
\begin{tabular}{|c|c|c|}
\hline
\textbf{Method} & \textbf{PSNR} & \textbf{SSIM}  \\
\hline
DeepDeblur {\color{green} \cite{multi_scale2017,github_deepdeblur}} & 32.89 & 0.9207\\
\hline
FMD-cGAN (ours)  & 31.79 & \textbf{0.9804}\\
\hline
\end{tabular}
\end{center}
\caption{The table shows the PSNR and SSIM comparison between FMD-cGAN (ours) and DeepDeblur {\color{green} \cite{multi_scale2017,github_deepdeblur}} on the REDS test dataset.}
\label{tab:Performance and efficiency comparison on the REDS test dataset}
\end{minipage}\hspace*{0.75cm}%
\begin{minipage}{0.5\textwidth}
\begin{center}
\begin{tabular}{|*4{c|}}
\hline
\textbf{Model} & \textbf{Dataset} & \textbf{\#Train Images} & \textbf{\#Test Images} \\
\hline
FMD-cGAN$_{WILD}$ & GoPro & 2103 & 1111 \\
\hline
FMD-cGAN$_{Comb}$ & 1. REDS & 24000 & 3000 \\
\cline{2-4}
 & 2. GoPro & 2103 & 1111 \\
\hline
\end{tabular}
\end{center}
\caption{The table summarises training details of our methods.}
\label{tab:training details}
\end{minipage}
\vspace{-9mm}
\end{table}

\section{Training Details}
\label{sec:training_details}
The Pytorch\footnote{https://pytorch.org/} deep learning library is used to implement our model. The training of the model is accomplished on a single Nvidia Quadro RTX 5000 GPU using different datasets. The model takes image patches as input and fully convolutional to be used on images of arbitrary size. There is no change in the learning rate for the first 150 epochs; after it, we decrease the learning rate linearly to zero for the subsequent 150 epochs. We used Adam {\color{green} \cite{adam}} optimizers for loss functions in both the Generator and the Discriminator with a learning rate of 0.0001. During the training time, we kept the batch size of 1, which gives a better result.\\
Furthermore, we used the dropout layer (rate=0) and the Instancenormalization layer instead of the batch-normalization layer concept both for the Generator and the Discriminator {\color{green} \cite{pix2pix}}. The training time of the network is approximately 2.5 days, which is significantly less than its competitive network. We have provided training details in Table {\color{red} \ref{tab:training details}}. We discuss the two variants of FMD-cGAN as follows. \\

\noindent \textbf{(1). FMD-cGAN$_{wild}$}: our first trained model is \textbf{WILD}, which represents that the model is trained only on a single dataset such as GoPro and REDS dataset on which we are going to evaluate it. For example, in the case of the GoPro dataset model is trained on 2103 pairs of blur and sharp images of the GoPro dataset. \\

\noindent \textbf{(2) FMD-cGAN$_{comb}$}: \label{fmdcgan_comb} The second trained model is \textbf{Comb}, which is first trained on the REDS training dataset; after training, we evaluate its performance on the REDS testing dataset. Now we train this pre-trained model on the GoPro dataset. We test both trained models \textbf{Comb} and \textbf{WILD} final performance on the GoPro dataset's 1111 test images.

\section{Experimental Results}
\label{sec:experimentalResults}
 We compare the results of our FMD-cGAN with relevant models using the standard performance metrics (PSNR, SSIM). We also show inference time of each model (i.e., average running time per image) on a single \textbf{GPU  (Nvidia RTX 5000)} and \textbf{CPU (2 X Intel Xeon 4216 (16C))}. To calculate Number of parameters and Number of MACs operations in PyTorch based model, we use pytorch-summary\footnote{https://github.com/sksq96/pytorch-summary} and torchprofile\footnote{https://github.com/zhijian-liu/torchprofile} libraries.

\begin{table}[!ht]
\renewcommand{\arraystretch}{1.2}
\begin{minipage}{0.45\textwidth}
\begin{center}
\begin{tabular}{|c|c|c|c|c|c|}
\hline
\textbf{\#ngf} & \textbf{PSNR} & \textbf{SSIM} & \textbf{Time (CPU)} & \textbf{\#Param} & \textbf{MACs}  \\
\hline
48 & 27.95 & 0.960 & 0.20 sec & \textbf{1.13M} & \textbf{10.60G}\\
\hline
\textbf{64} & 28.33 & 0.963 & 0.28 sec & 1.98M & 18.36G\\
\hline
96 & \textbf{28.52} & \textbf{0.964} & 0.5 sec & 4.41M & 40.23G\\
\hline
\end{tabular}
\end{center}
\caption{Performance and efficiency comparison on the different no. of generator filters (\#ngf)}
\label{tab:different generator frames}
\end{minipage}\hspace*{0.75cm}%
\begin{minipage}{0.5\textwidth}
\begin{center}
\begin{tabular}{|p{100pt}|c|c|c|c|}
\hline
\textbf{Model} & \textbf{PSNR} & \textbf{SSIM} & \textbf{\#Param} & \textbf{MACs}\\
\hline
Only ResNetBlock & \textbf{28.33} & 0.963 & 1.98M & 18.36G\\
\hline
Downsample + ResNetBlock & 28.24 & 0.962 & \textbf{1.661M} & 16.81G\\
\hline
Upsample + ResNetBlock & 28.19 & 0.961 & 1.663M & \textbf{11.79G}\\
\hline
\end{tabular}
\end{center}
\caption{Performance comparison after applying convolution decomposition in different parts of network and \#ngf=64}
\label{tab:convolution decomposition in different parts}
\end{minipage}
\vspace{-8mm}
\end{table}

\begin{table}[!ht]
        \centering
        \begin{tabular}{cm{45mm}m{45mm}m{45mm}}
           \toprule
            \textbf{Model} & \textbf{Example 1} & \textbf{Example 2} & \textbf{Example 3} \\
            \midrule
            Blurry &  \includegraphics[width=\linewidth,height=30mm]{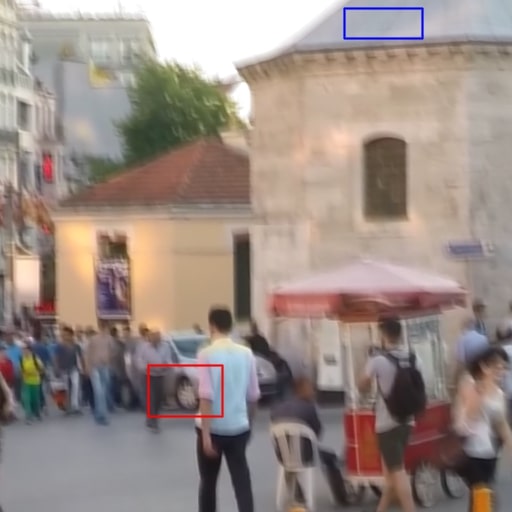} &
            \includegraphics[width=\linewidth,height=30mm]{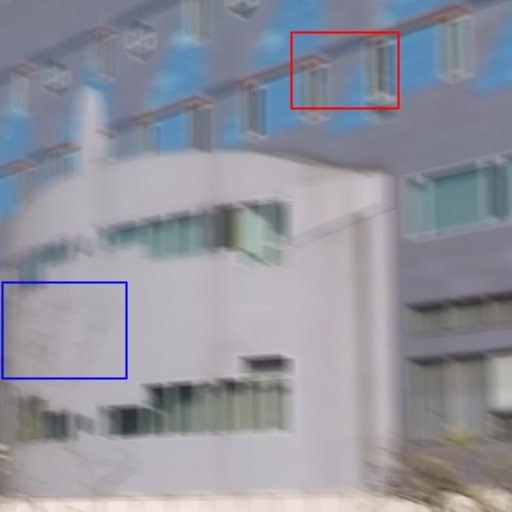} &
            \includegraphics[width=\linewidth,height=30mm]{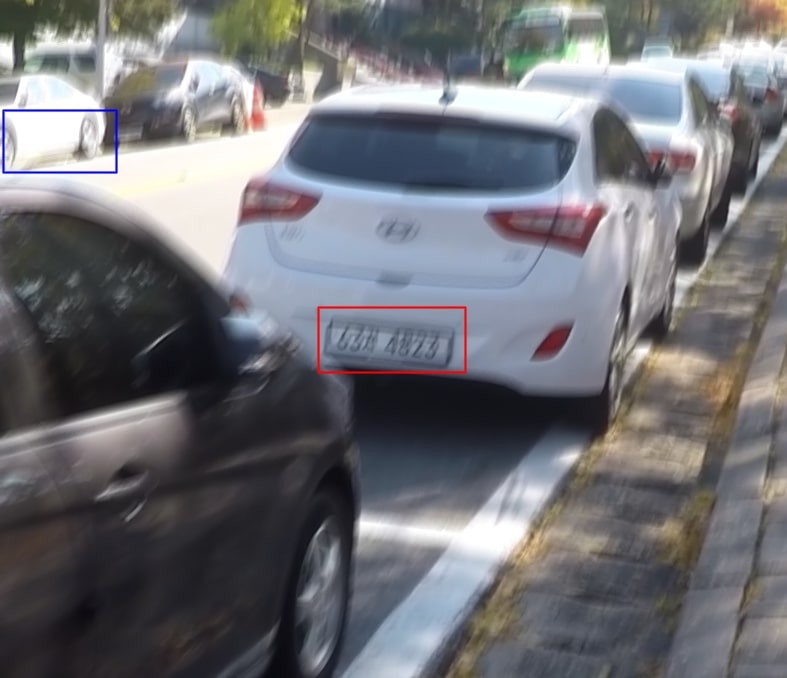} \\
            \pbox{25cm}{DeblurGANv2 \\ {\color{green} \cite{deblur_v2}}} &  \includegraphics[width=\linewidth,height=30mm]{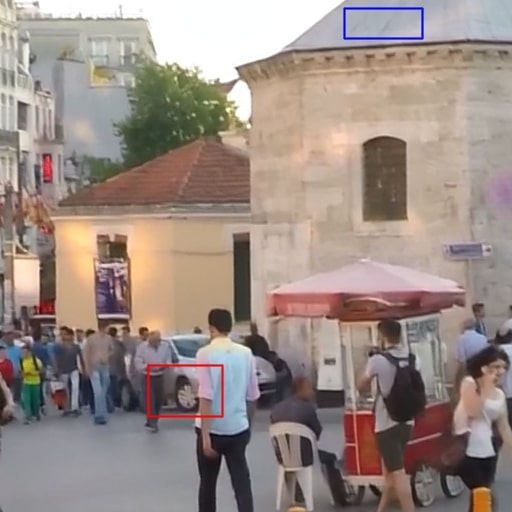} &
            \includegraphics[width=\linewidth,height=30mm]{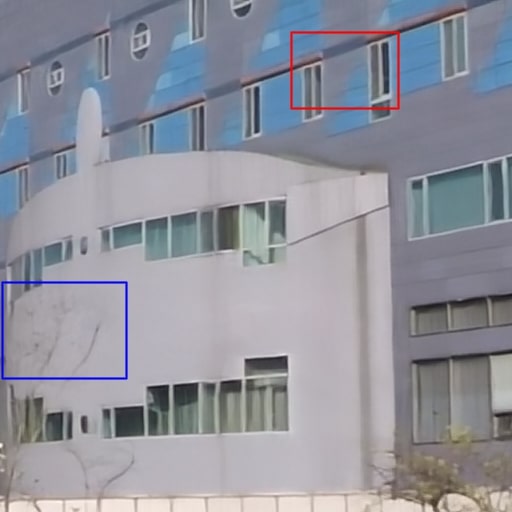} &
            \includegraphics[width=\linewidth,height=30mm]{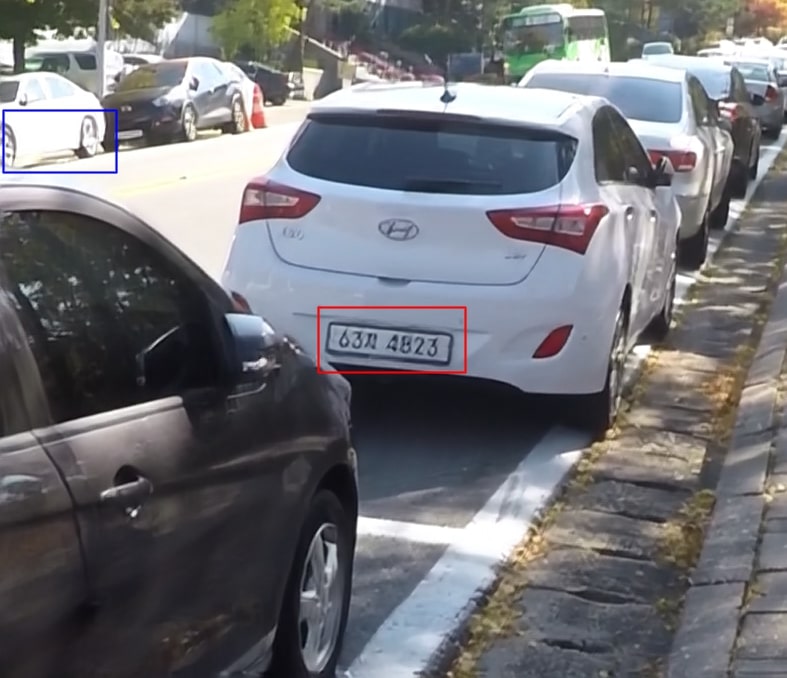} \\
            \pbox{20cm}{DeepDeblur \\ {\color{green} \cite{multi_scale2017,github_deepdeblur}}} &  \includegraphics[width=\linewidth,height=30mm]{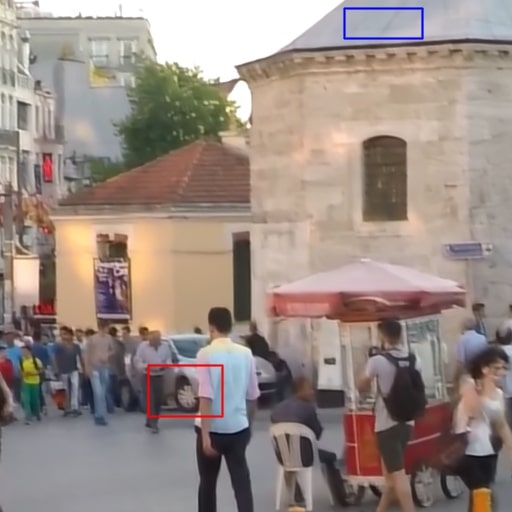} &
            \includegraphics[width=\linewidth,height=30mm]{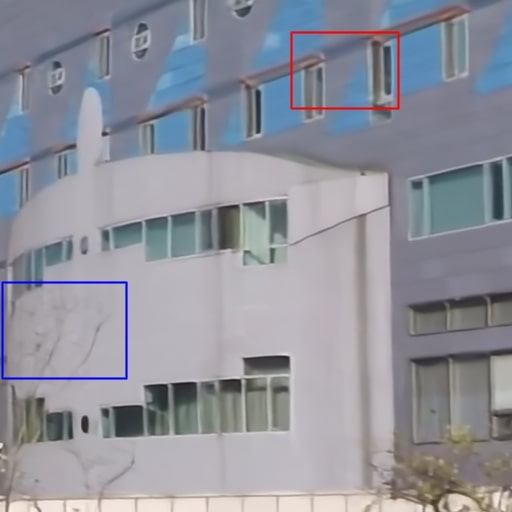} &
            \includegraphics[width=\linewidth,height=30mm]{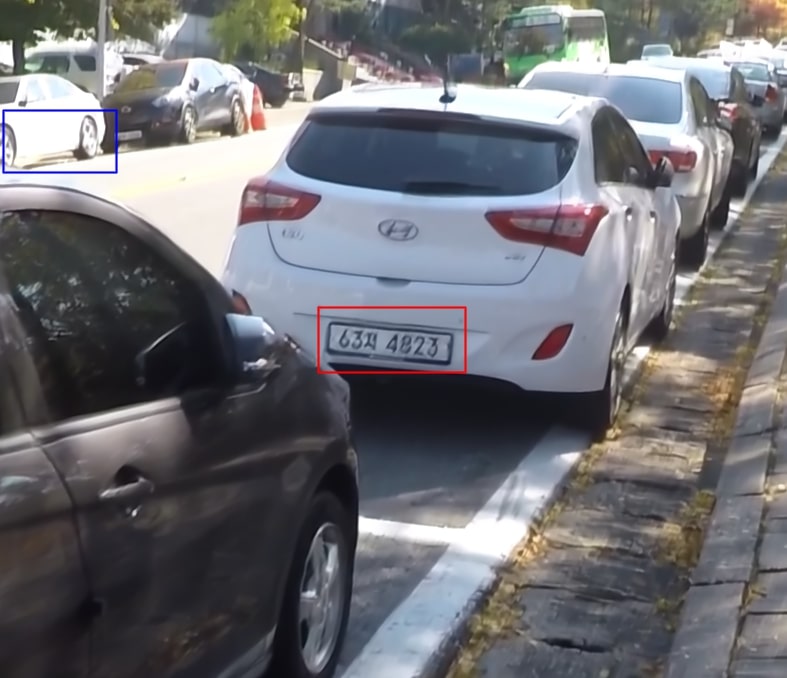} \\
            \pbox{20cm}{SRN \\ {\color{green} \cite{SRN}}} & 
            \includegraphics[width=\linewidth,height=30mm]{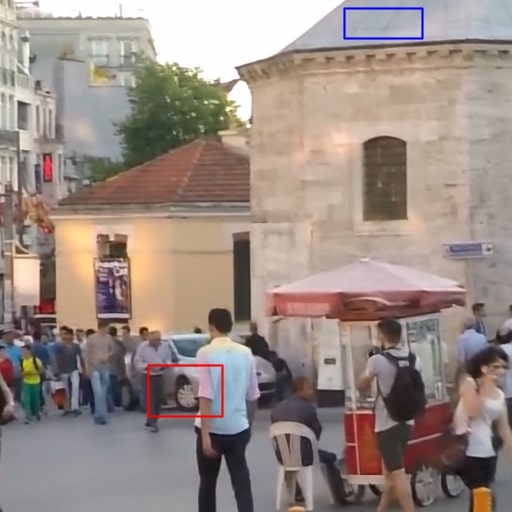} &
            \includegraphics[width=\linewidth,height=30mm]{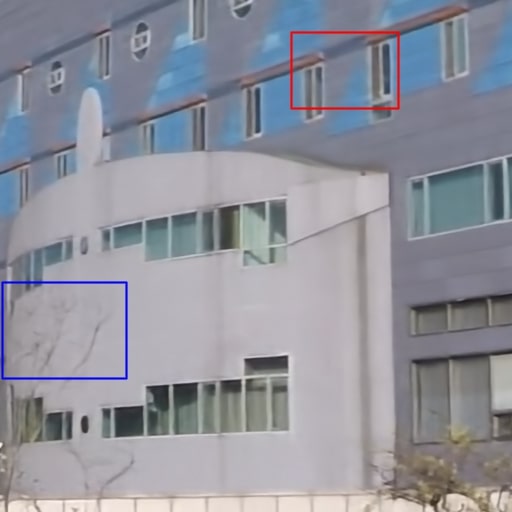} &
            \includegraphics[width=\linewidth,height=30mm]{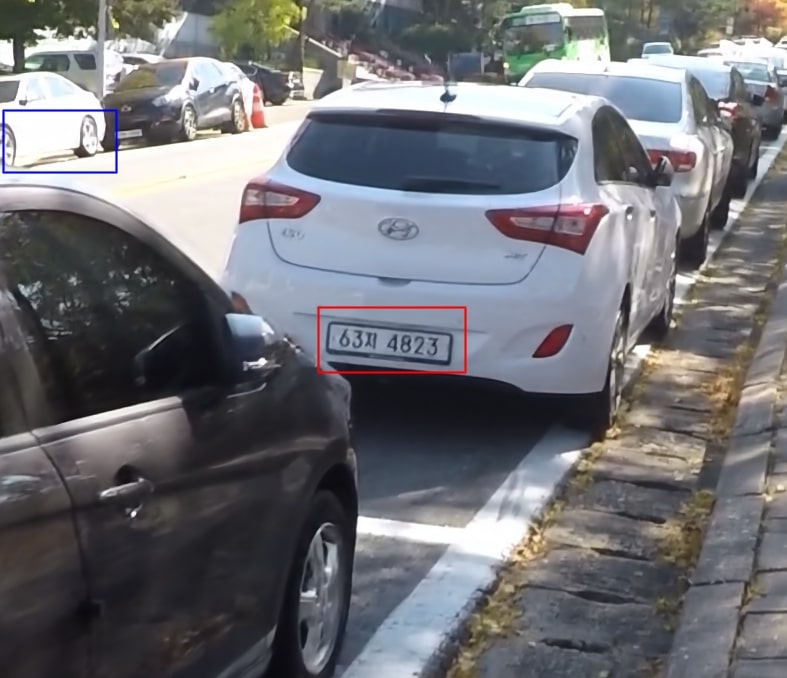} \\
            \pbox{20cm}{FMD-cGAN$_{Wild}$\\(Ours)} &  \includegraphics[width=\linewidth,height=30mm]{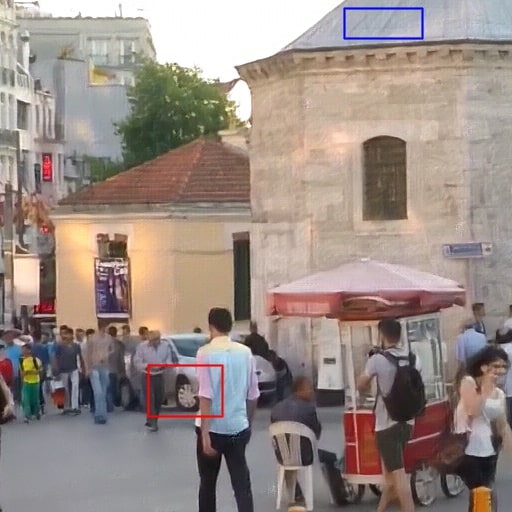} &
            \includegraphics[width=\linewidth,height=30mm]{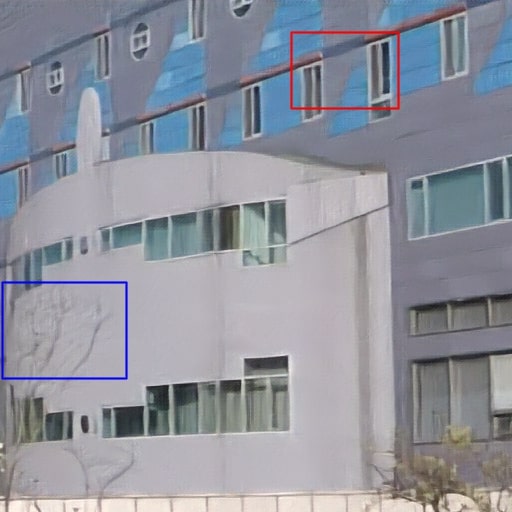} &
            \includegraphics[width=\linewidth,height=30mm]{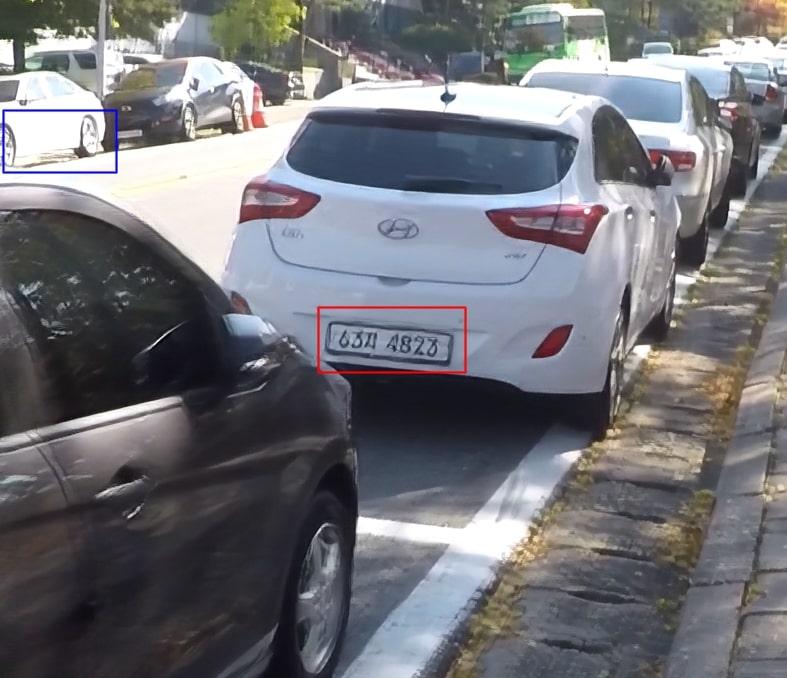} \\
            \pbox{20cm}{FMD-cGAN$_{Comb}$\\(Ours)} &  \includegraphics[width=\linewidth,height=30mm]{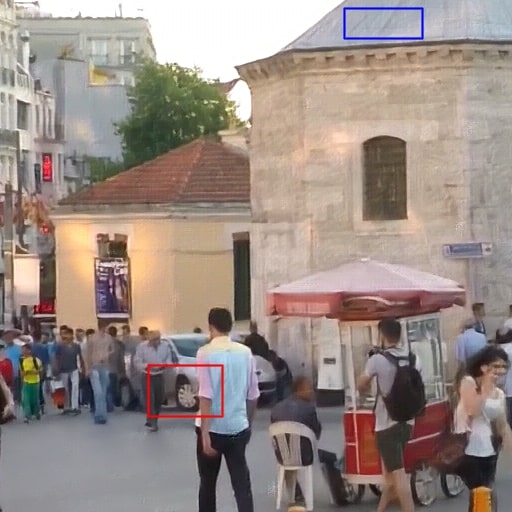} &
            \includegraphics[width=\linewidth,height=30mm]{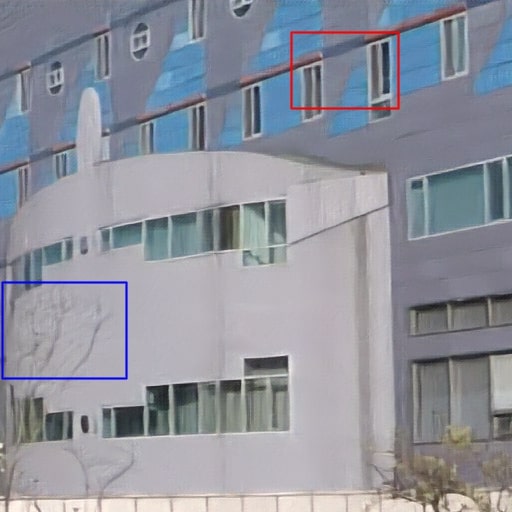} &
            \includegraphics[width=\linewidth,height=30mm]{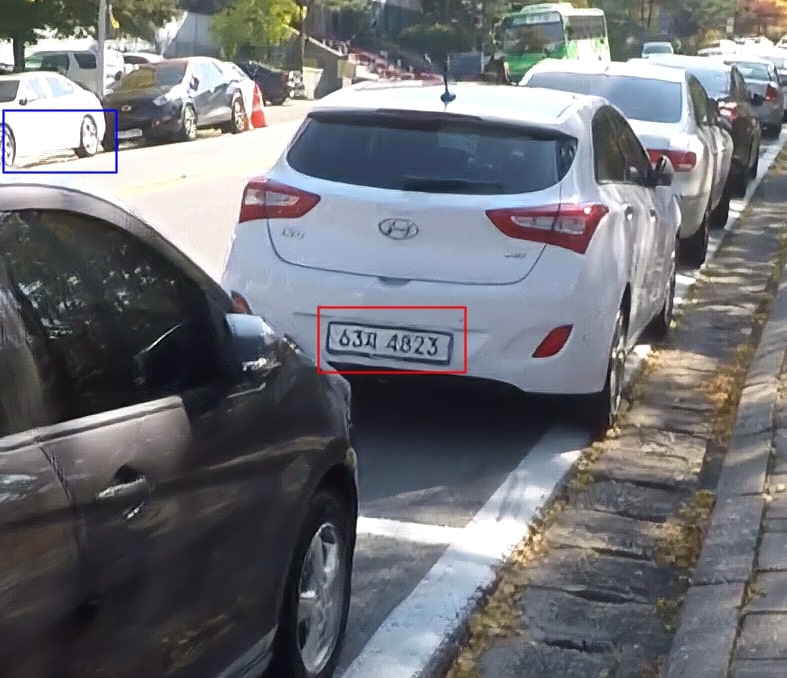} \\
            \bottomrule
        \end{tabular}
        \captionof{figure}{The figure shows visual comparison on the GoPro dataset (images are best viewed after zooming).}
        \label{tbl:comparison on the GoPro dataset}
        \vspace{-10mm}
\end{table}

\subsection{Quantitative Evaluation on GoPro Dataset}
Here, we discuss the performance of our method on GoPro Dataset. We used 1111 pairs of blur and sharp images from GoPro test dataset for evaluation. We compare our model's results with other state-of-the-art model's results: where Sun et al. {\color{green} \cite{nonuniform2015blur}} is a traditional method, while others are deep learning-based methods: Xu et al. {\color{green} \cite{unnaturalsparse}}, DeepDeblur {\color{green} \cite{multi_scale2017}}, DeepFilter {\color{green} \cite{ramakrishnan}}, $DeblurGAN$ {\color{green} \cite{deblurgan}}, $DeblurGANv2$ {\color{green} \cite{deblur_v2}} and SRN {\color{green} \cite{SRN}}. We use PSNR and SSIM value of other methods from their respective papers. 

We show the results in \textbf{Table \color{red}\ref{tab:Performance and efficiency comparison on the GoPro test dataset}}. It could be observed that FMD-cGAN (ours) has high efficiency in terms of performance and inference time. FMD-cGAN also has the lowest inference time, and in terms of no. of parameters and macs operations also has the lowest value. Furthermore, FMD-cGAN output PSNR and SSIM values comparable to the other models in comparison.

\subsection{Quantitative Evaluation on REDS Dataset}
We also show the performance of our framework on the REDS dataset. We used 3000 pairs of blur and sharp images from REDS test dataset for evaluation. We compare the performance of FMD-cGAN (ours) with the DeepDeblur model {\color{green}\cite{multi_scale2017}}. We used the results of DeepDeblur from official GitHub repository - DeepDeblur-PyTorch\footnote{https://github.com/SeungjunNah/DeepDeblur-PyTorch}. 

We show the results in \textbf{Table {\color{red}\ref{tab:Performance and efficiency comparison on the REDS test dataset}}}. It could be observed that our method achieves high SSIM and PSNR values which are comparable to DeepDeblur {\color{green} \cite{multi_scale2017}}. We emphasise that our network has a significantly lesser size as compared to DeepDeblur {\color{green} \cite{multi_scale2017}}. Currently, only the DeepDeblur model used the REDS dataset for training and performance evaluation.

\subsection{Visual Comparison}
Fig. {\color{red} \ref{tbl:comparison on the REDS dataset} } shows the visual comparison on the REDS dataset. It could be observed that FMD-cGAN (ours) restore images comparable to the relevant top-performing works such as DeepDeblur {\color{green} \cite{multi_scale2017}} and SRN {\color{green} \cite{SRN}}. For example, row 1 of Fig. {\color{red} \ref{tbl:comparison on the REDS dataset} } shows that our method preserves the fine object structure details (i.e., building) which are missing in the blurry image. \\
Fig. {\color{red} \ref{tbl:comparison on the GoPro dataset}} shows the visual comparison results on the GoPro dataset. It could be observed that the output of our method is visually appealing in the presence of motion blur in the input image (see Example~3 of Fig. {\color{red} \ref{tbl:comparison on the GoPro dataset}}). To provide more clarity, we show the results for both FMD-cGAN$_{Wild}$ and FMD-cGAN$_{Comb}$ (Sec. {\color{red} \ref{sec:training_details}}). FMD-cGAN (ours) is faster and output better reconstruction than other motion deblurring methods even though our model has fewer parameters (Table {\color{red} \ref{tab:Performance and efficiency comparison on the GoPro test dataset}}). We have provided the extended versions of Fig. {\color{red} \ref{tbl:comparison on the REDS dataset} } and Fig. {\color{red} \ref{tbl:comparison on the GoPro dataset}} in the supplementary material for better visual comparisons. 

\section{Ablation Study}
\label{sec:ablation_study}
Table {\color{red} \ref{tab:different generator frames}} shows an ablation study on the generator network architecture for  different design choices. Here, we train and test our network's performance only on the GoPro dataset. Suppose \#ngf denotes the initial layer's filters count in the generator network, affecting filters count of subsequent layers. Table {\color{red} \ref{tab:different generator frames}} demonstrates how \#ngf affects model performance. It could be observed that if we increase the \#ngf then image quality (PSNR) will increase. However, it increases \#parameters and MACs operations also, affecting inference time and model size.

We divide our generator network into three parts according to its structure: Downsample (two 3x3 convolutions), ResnetBlocks (9 blocks), and Upsample (two 3x3 deconvolutions). To check the network performance, we put separable convolution into different parts. Table {\color{red} \ref{tab:convolution decomposition in different parts}} demonstrates model performance after applying convolution decomposition in different parts of the generator network. ResNet blocks do most of the computation in the network; from Table~{\color{red} \ref{tab:convolution decomposition in different parts}}, we can see applying convolution decomposition in this part giving better performance.  

\section{Conclusion}
We proposed a Fast Motion Deblurring method (FMD-cGAN) for a single image. FMD-cGAN does not require knowledge of the blur kernel. Our method uses the conditional generative adversarial network for this task and is optimized using the multi-part loss function.  Our method shows that using MobileNetv1 architecture consists of depthwise separable convolution to reduce computational cost and memory requirement without losing accuracy. We also proposed that using Hinge loss in the network gives good results. Our method produces better blur-free images, as confirmed by the quantitative and visual comparisons. FMD-cGAN is faster with low inference time and memory requirements, and it outperforms various state-of-the-art models for blind motion deblurring of a single image (\textbf{Table \color{red}\ref{tab:Performance and efficiency comparison on the GoPro test dataset}}). We propose as future work to deploy our model in lightweight devices for real-time image deblurring tasks.

%
%

%
%
%
%

\clearpage

\begin{center}
    {\Huge \textbf{FMD-cGAN: Fast Motion Deblurring using cGAN \\(Supplementary Material)}} \\~\\
    {\large Submission Id 201}\\~\\~\\
\end{center}

In the supplementary material, we provide the extended version of the figures for better visual comparisons. ~\\
\begin{figure*}[!h]
    \centering
    \begin{minipage}{0.45\linewidth}\centering
    \includegraphics[width=\linewidth]{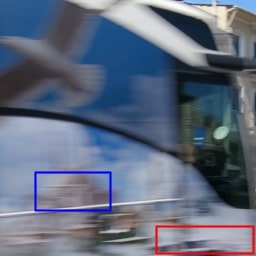} (a) Blurry
    \end{minipage}\hspace{0.25cm} %
    \begin{minipage}{0.45\linewidth}\centering
    \includegraphics[width=\linewidth]{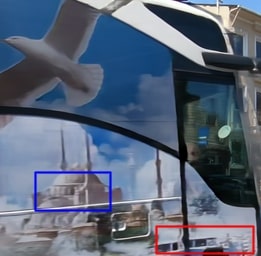} (b) DeepDeblur {\color{green} \cite{multi_scale2017,github_deepdeblur}} 
    \end{minipage}\\%
        \begin{minipage}{0.45\linewidth}\centering
    \includegraphics[width=\linewidth]{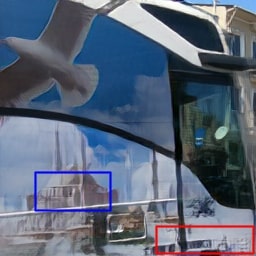} (c) FMD-cGAN (Ours)
    \end{minipage}\hspace{0.25cm} %
    \begin{minipage}{0.45\linewidth}\centering
    \includegraphics[width=\linewidth]{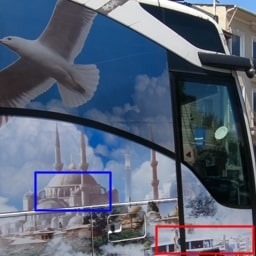} (d) Sharp 
    \end{minipage}%
    \caption{visual comparison on the REDS dataset}
    \label{fig:example1_reds}
\end{figure*}


\begin{figure*}[!h]
    \centering
    \begin{minipage}{0.45\linewidth}\centering
    \includegraphics[width=\linewidth]{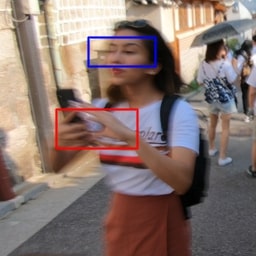} (a) Blurry
    \end{minipage}\hspace{0.25cm} %
    \begin{minipage}{0.45\linewidth}\centering
    \includegraphics[width=\linewidth]{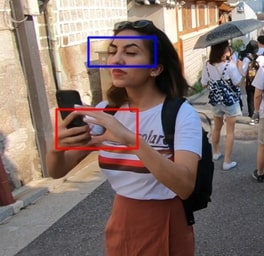} (b) DeepDeblur {\color{green} \cite{multi_scale2017,github_deepdeblur}}  
    \end{minipage}\\%
        \begin{minipage}{0.45\linewidth}\centering
    \includegraphics[width=\linewidth]{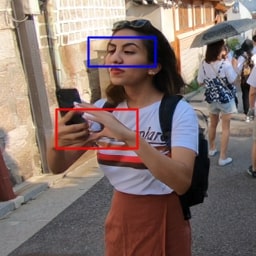} (c) FMD-cGAN (Ours)
    \end{minipage}\hspace{0.25cm} %
    \begin{minipage}{0.45\linewidth}\centering
    \includegraphics[width=\linewidth]{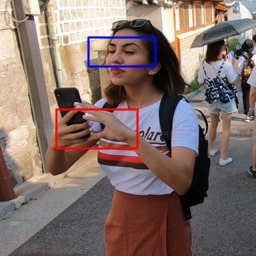} (d) Sharp 
    \end{minipage}%
    \caption{visual comparison on the REDS dataset}
    \label{fig:example2_reds}
\end{figure*}
\clearpage 


\begin{figure*}[!h]
    \centering
    \begin{minipage}{0.45\linewidth}\centering
    \includegraphics[width=\linewidth]{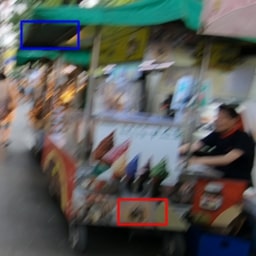} (a) Blurry
    \end{minipage}\hspace{0.25cm} %
    \begin{minipage}{0.45\linewidth}\centering
    \includegraphics[width=\linewidth]{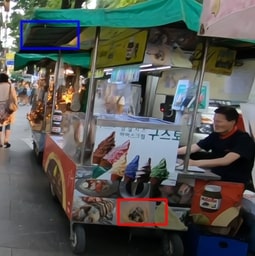} (b) DeepDeblur {\color{green} \cite{multi_scale2017,github_deepdeblur}} 
    \end{minipage}\\%
        \begin{minipage}{0.45\linewidth}\centering
    \includegraphics[width=\linewidth]{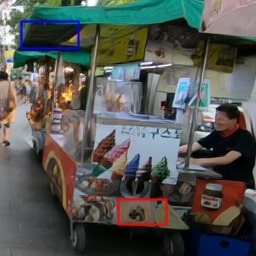} (c) FMD-cGAN (Ours)
    \end{minipage}\hspace{0.25cm} %
    \begin{minipage}{0.45\linewidth}\centering
    \includegraphics[width=\linewidth]{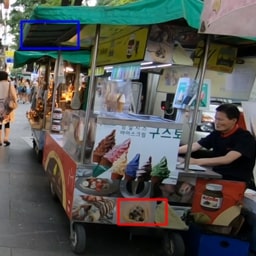} (d) Sharp 
    \end{minipage}%
    \caption{visual comparison on the REDS dataset}
    \label{fig:example3_reds}
\end{figure*}


\begin{figure*}[!h]
    \centering
    \begin{minipage}{0.45\linewidth}\centering
    \includegraphics[width=\linewidth]{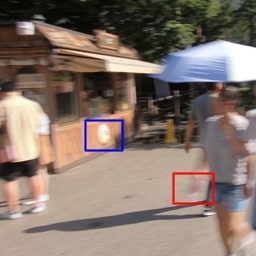} (a) Blurry
    \end{minipage}\hspace{0.25cm} %
    \begin{minipage}{0.45\linewidth}\centering
    \includegraphics[width=\linewidth]{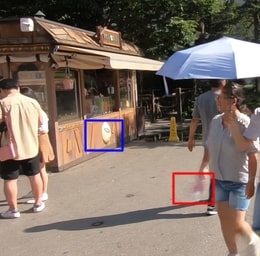} (b) DeepDeblur {\color{green} \cite{multi_scale2017,github_deepdeblur}}  
    \end{minipage}\\%
        \begin{minipage}{0.45\linewidth}\centering
    \includegraphics[width=\linewidth]{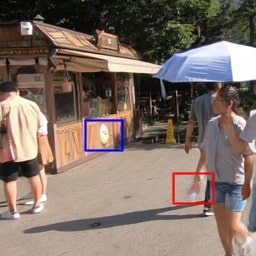} (c) FMD-cGAN (Ours)
    \end{minipage}\hspace{0.25cm} %
    \begin{minipage}{0.45\linewidth}\centering
    \includegraphics[width=\linewidth]{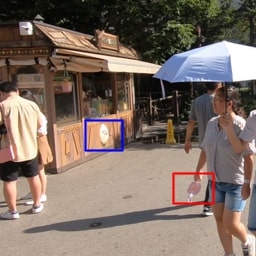} (d) Sharp 
    \end{minipage}%
    \caption{visual comparison on the REDS dataset}
    \label{fig:example4_reds}
\end{figure*}
\clearpage 

\begin{figure*}
    \centering
    \begin{minipage}{0.4\linewidth}\centering
    \includegraphics[width=\linewidth]{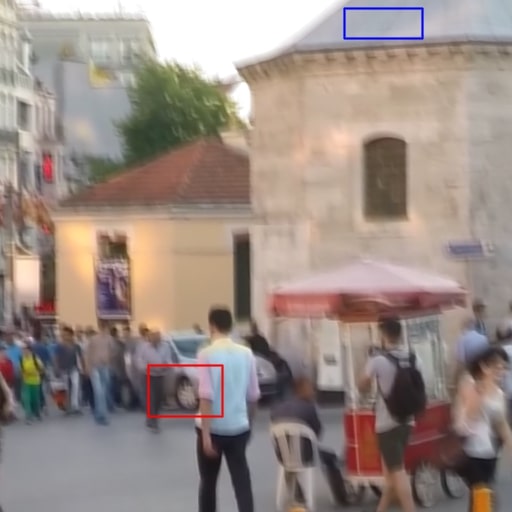} (a) Blurry
    \end{minipage}\hspace{0.25cm} %
    \begin{minipage}{0.4\linewidth}\centering
    \includegraphics[width=\linewidth]{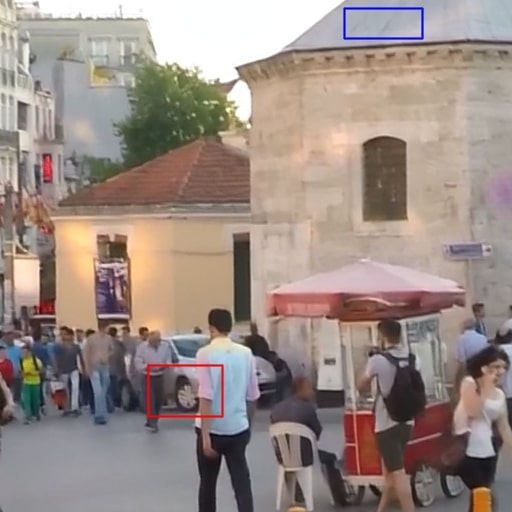} (b) DeblurGANv2 {\color{green} \cite{deblur_v2}} 
    \end{minipage}\\%
        \begin{minipage}{0.4\linewidth}\centering
    \includegraphics[width=\linewidth]{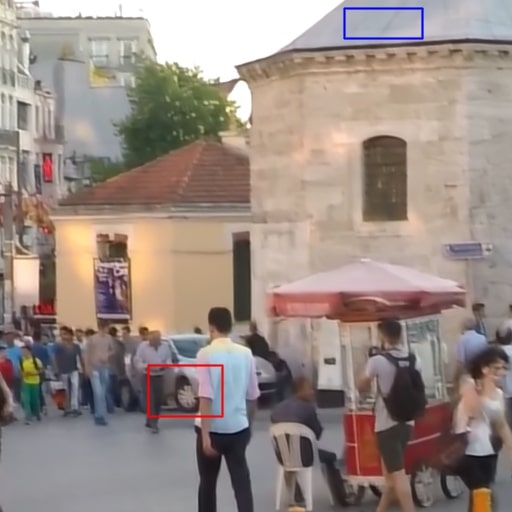} (c) DeepDeblur {\color{green} \cite{multi_scale2017,github_deepdeblur}}
    \end{minipage}\hspace{0.25cm} %
    \begin{minipage}{0.4\linewidth}\centering
    \includegraphics[width=\linewidth]{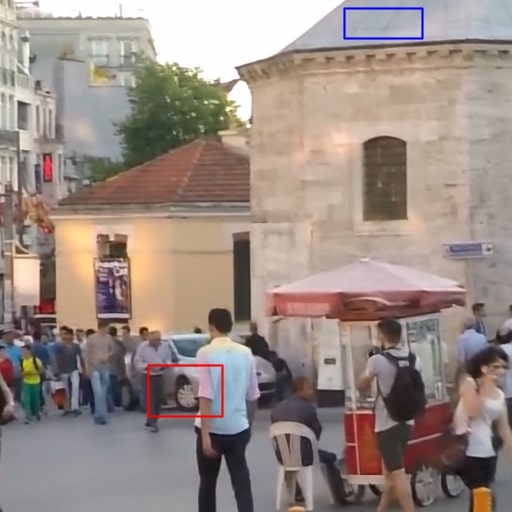} (d) SRN {\color{green} \cite{SRN}} 
    \end{minipage}\\%
        \begin{minipage}{0.4\linewidth}\centering
    \includegraphics[width=\linewidth]{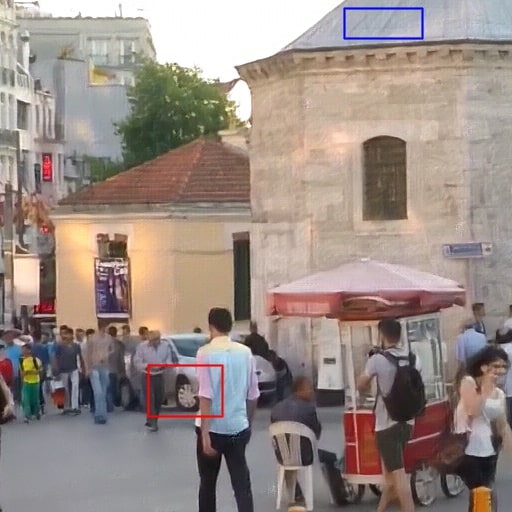} (e) FMD-cGAN$_{Wild}$ (Ours)
    \end{minipage}\hspace{0.25cm} %
    \begin{minipage}{0.4\linewidth}\centering
    \includegraphics[width=\linewidth]{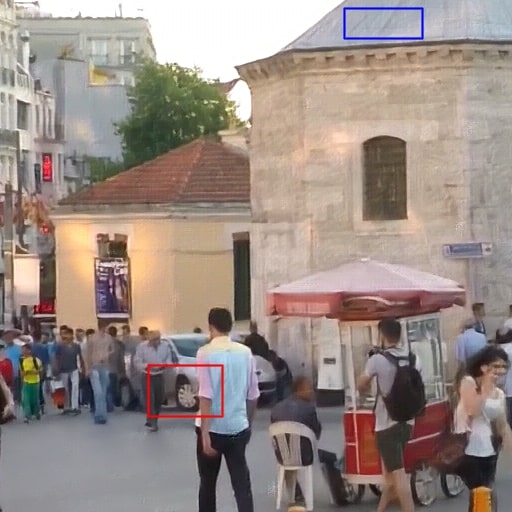} (f) FMD-cGAN$_{Comb}$ (Ours) 
    \end{minipage}
    \caption{Visual comparison on the GoPro dataset.}
    \label{fig:example1_gopro}
\end{figure*}


\begin{figure*}
    \centering
    \begin{minipage}{0.4\linewidth}\centering
    \includegraphics[width=\linewidth]{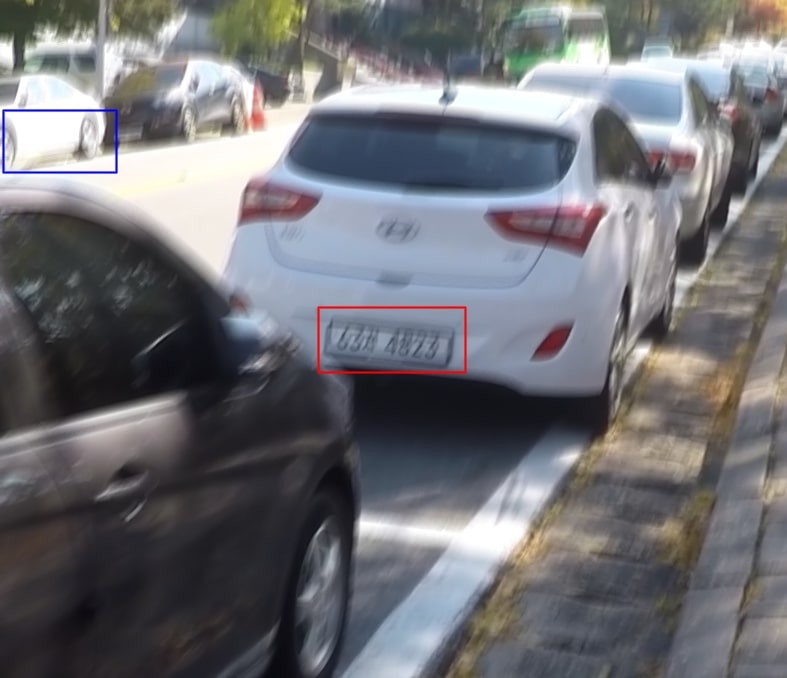} (a) Blurry
    \end{minipage}\hspace{0.25cm} %
    \begin{minipage}{0.4\linewidth}\centering
    \includegraphics[width=\linewidth]{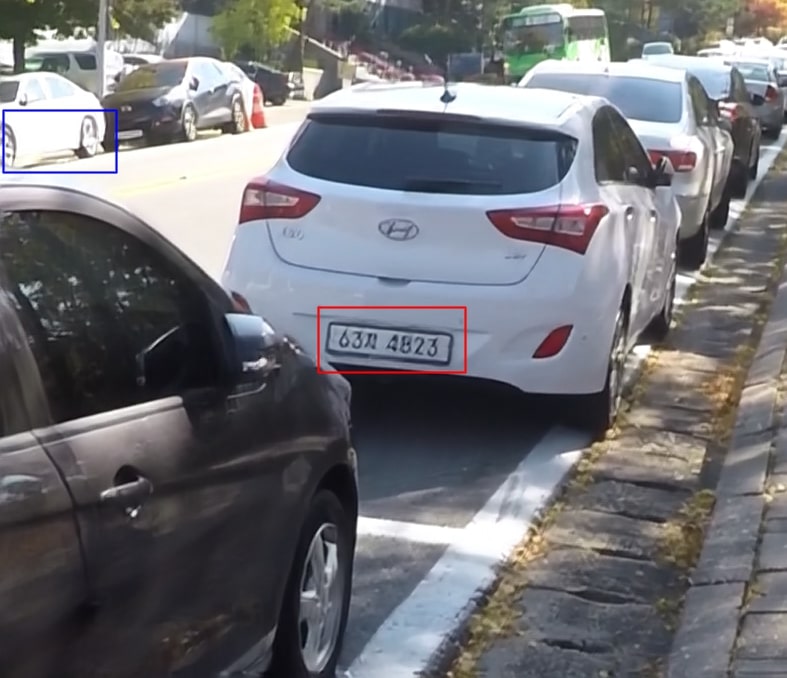} (b) DeblurGANv2 {\color{green} \cite{deblur_v2}}  
    \end{minipage}\\%
        \begin{minipage}{0.4\linewidth}\centering
    \includegraphics[width=\linewidth]{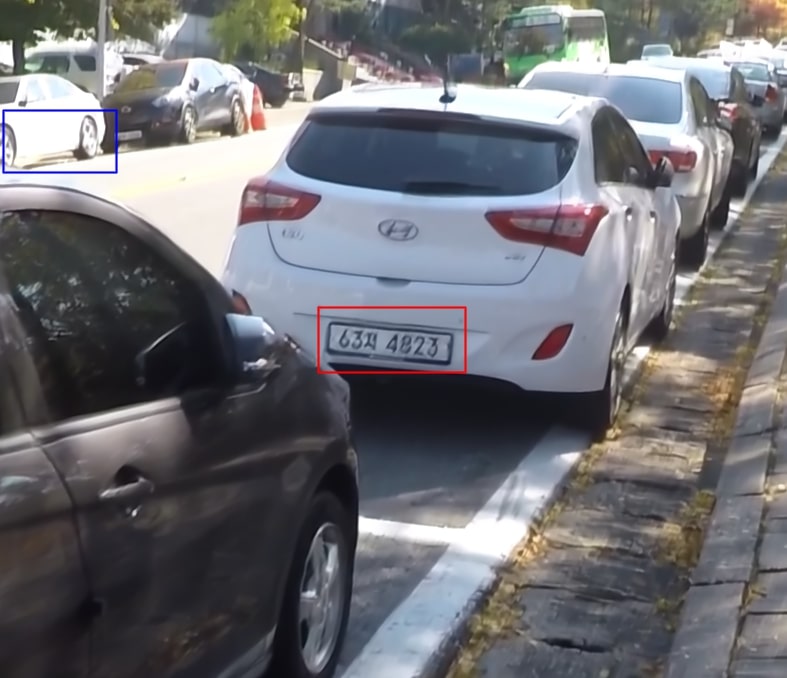} (c) DeepDeblur {\color{green} \cite{multi_scale2017,github_deepdeblur}}
    \end{minipage}\hspace{0.25cm} %
    \begin{minipage}{0.4\linewidth}\centering
    \includegraphics[width=\linewidth]{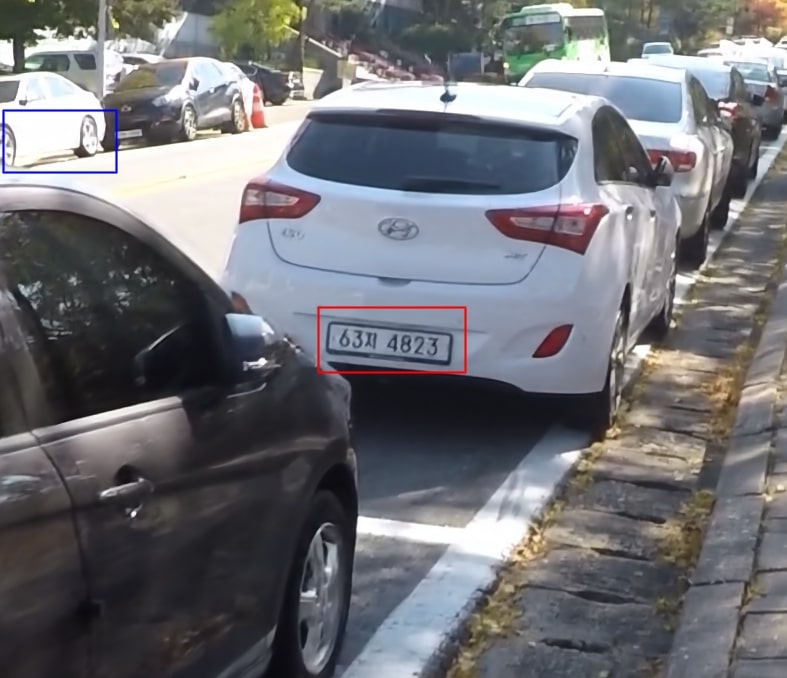} (d) SRN {\color{green} \cite{SRN}} 
    \end{minipage}\\%
        \begin{minipage}{0.4\linewidth}\centering
    \includegraphics[width=\linewidth]{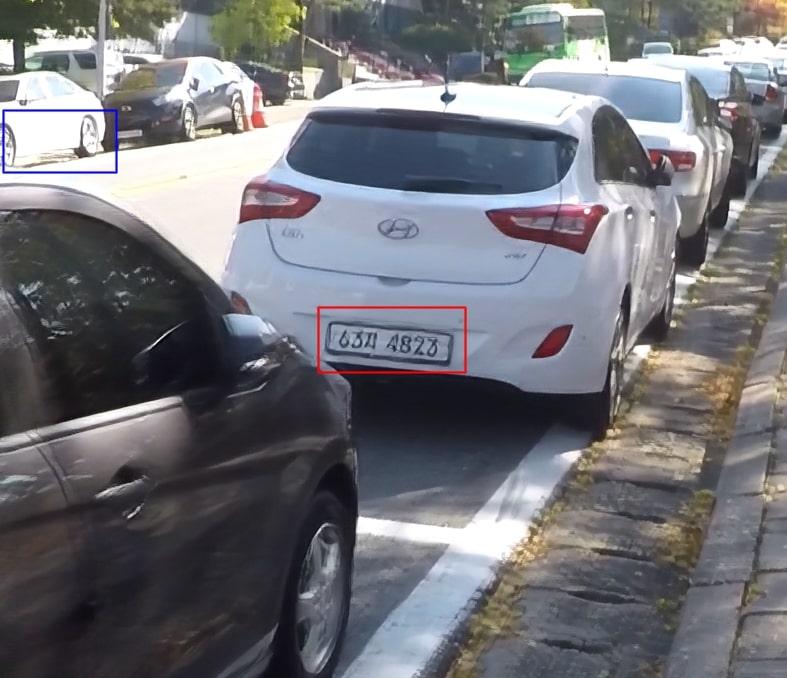} (e) FMD-cGAN$_{Wild}$ (Ours)
    \end{minipage}\hspace{0.25cm} %
    \begin{minipage}{0.4\linewidth}\centering
    \includegraphics[width=\linewidth]{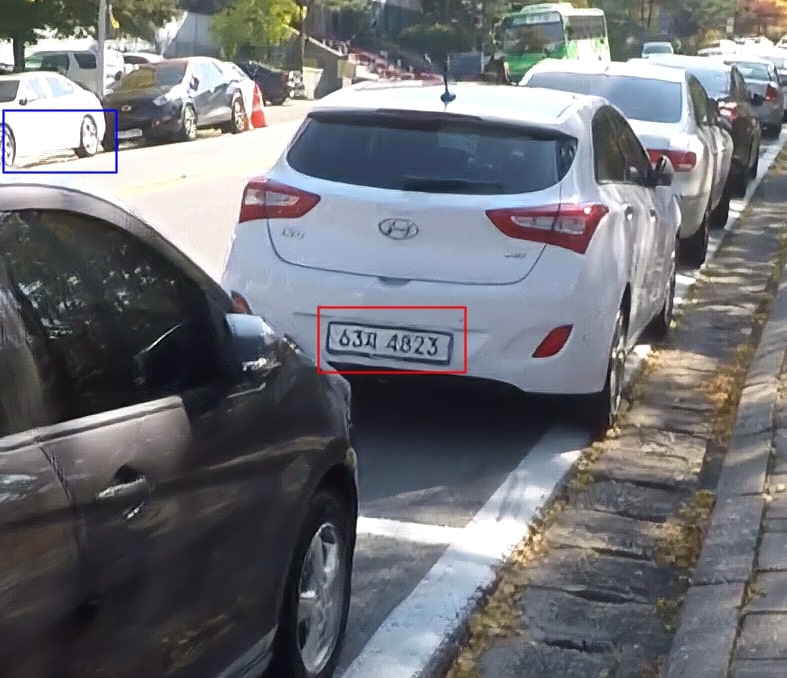} (f) FMD-cGAN$_{Comb}$ (Ours)
    \end{minipage}
    \caption{Visual comparison on the GoPro dataset.}
    \label{fig:example2_gopro}
\end{figure*}


\begin{figure*}
    \centering
    \begin{minipage}{0.4\linewidth}\centering
    \includegraphics[width=\linewidth]{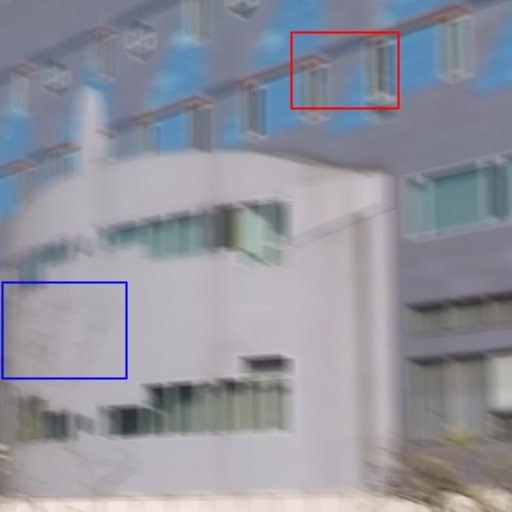} (a) Blurry
    \end{minipage}\hspace{0.25cm} %
    \begin{minipage}{0.4\linewidth}\centering
    \includegraphics[width=\linewidth]{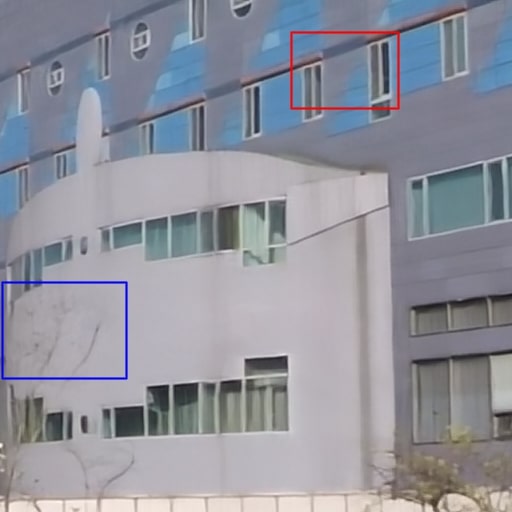} (b) DeblurGANv2 {\color{green} \cite{deblur_v2}}  
    \end{minipage}\\%
        \begin{minipage}{0.4\linewidth}\centering
    \includegraphics[width=\linewidth]{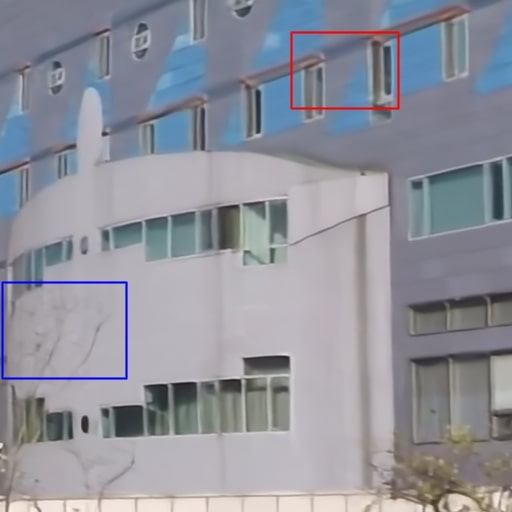} (c) DeepDeblur {\color{green} \cite{multi_scale2017,github_deepdeblur}}
    \end{minipage}\hspace{0.25cm} %
    \begin{minipage}{0.4\linewidth}\centering
    \includegraphics[width=\linewidth]{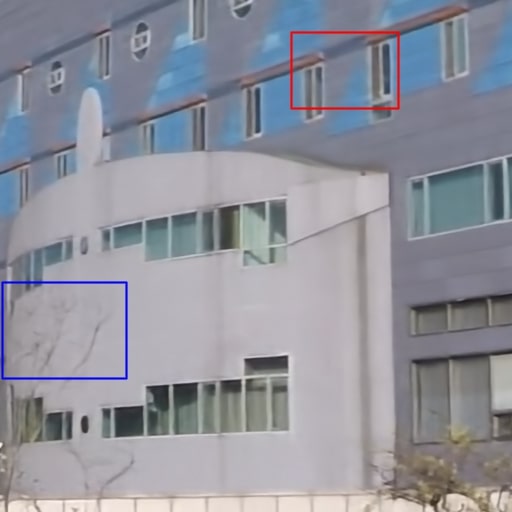} (d) SRN {\color{green} \cite{SRN}}  
    \end{minipage}\\%
        \begin{minipage}{0.4\linewidth}\centering
    \includegraphics[width=\linewidth]{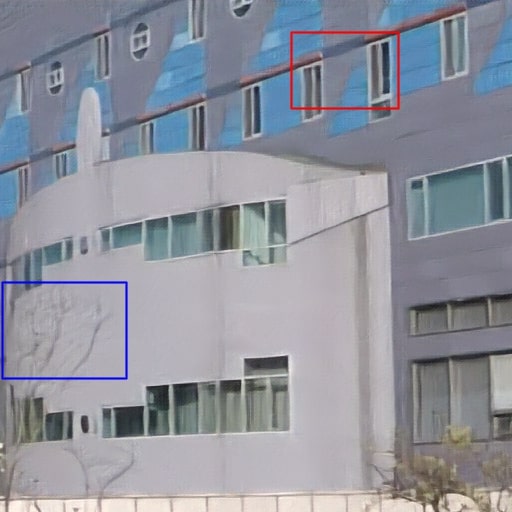} (e) FMD-cGAN$_{Wild}$ (Ours)
    \end{minipage}\hspace{0.25cm} %
    \begin{minipage}{0.4\linewidth}\centering
    \includegraphics[width=\linewidth]{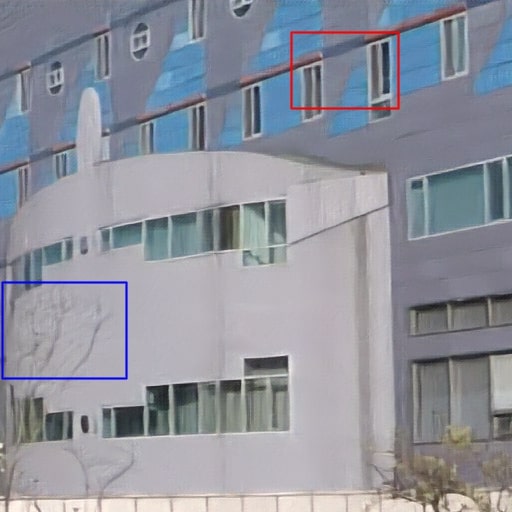} (f) FMD-cGAN$_{Comb}$ (Ours) 
    \end{minipage}
    \caption{Visual comparison on the GoPro dataset.}
    \label{fig:example3_gopro}
\end{figure*}


\end{document}